\documentstyle[delarray,math-ecai96]{ecai96}

\sectionfoot{Natural Language Processing}
\authorfoot{Rolf Backofen}
\input{amssym}

\newcommand{\makermfunction}[1]{\mathop{\mathchoice{\mbox{\rm #1}}{\mbox{\rm #1}}{\mbox{\scriptsize\rm #1}}{\mbox{\scriptsize\rm #1}}}\nolimits}
\newcommand{\makeitfunction}[1]{\mathop{\mathchoice{\mbox{\it #1}}{\mbox{\it #1}}{\mbox{\scriptsize\it #1}}{\mbox{\scriptsize\it #1}}}\nolimits}

\newcommand{\pterms}{s}
\newcommand{\ptermt}{t}
\newcommand{\ptermp}{p}
\newcommand{\ptermq}{q}

\newcommand{\simplified}{simplified}

\newcommand{\control}{{\rm con}}

\newcommand{\controlconst}[1]{<^{\rm #1}}

\newcommand{\decfun}{{\rm DFun}}

\newcommand{\outgoing}[2]{\makermfunction{outgoing}_{#1}(#2)}

\newlength{\defindent}
\newcommand{\treesort}{\ifmmode{{\makeitfunction{tree}}}\else{{\it tree}}\fi}
\newcommand{\pathsort}{\ifmmode{{\makeitfunction{path}}}\else{{\it path}}\fi}

\def\F{{\ifmmode{{\rm F}}\else{${\rm F}$}\fi}}
\def\Finftheory{{\ifmmode{\theory{\infmod}}\else{$theory{\infmod}$}\fi}}
\def\Frattheory{{\ifmmode{\theory{\ratmod}}\else{$\theory{\ratmod}$}\fi}}

\def\makeatletter{\catcode`\@=11 }
\chardef\other=12
\def\makeatother{\catcode`\@=\other}

\makeatletter           
\newwrite\protokoll
\openout\protokoll=\jobname.tdx
\def\@begintheorem#1#2{\immediate\write\protokoll{\string\indexentry {#1 #2}{\thepage}}%
\it \trivlist \item[\hskip \labelsep{\bf #1\ #2}]}

\def\setrefto#1{\let\@tempa\protect%
\def\protect{\noexpand\protect\noexpand}%
\edef\@currentlabel{#1}%
\let\protect\@tempa}

\newcounter{@axiomctrrem}
\newcounter{@axiomctr}
\renewcommand{\the@axiomctr}{Ax\arabic{@axiomctr}}

\def\axitem{\@ifnextchar [{\@axitem}%
{\refstepcounter{@axiomctr}%
\setcounter{@axiomctrrem}{\value{@axiomctr}}%
\@axitem[\the@axiomctr]}}
\def\@axitem[#1]{\llap{(#1)} &\setrefto{#1}\ignorespaces}

{\end{tabular}\egroup\par}

\makeatother

\newcommand{\myalpha}{{{\mu}}}
\newcommand{\mybeta}{{{\nu}}}
\newcommand{\mydelta}{{{\delta}}}
\newcommand{\frel}[1]{{[#1]}}

\newcommand{\inftrees}{\frak{T}}
\newcommand{\rattrees}{\frak{R}}
\newcommand{\ratmod}{{\rattrees_{\F}}}
\newcommand{\infmod}{{\inftrees_{\F}}}

\newcommand{\Rsimpl}{{\cal R}_{\rm simpl}}

\newcommand{\Rpre}{{\cal R}_{\rm pre}}
\newcommand{\Rsolve}{{\cal R}_{\rm solve}}

\newcommand{\Thetasimpl}{\Theta_{\rm simpl}}

\newcommand{\basiccon}{\controlconst{{\rm basic}}}
\newcommand{\quasicon}{\controlconst{{\rm quasi}}}
\newcommand{\maxwell}{\controlconst{{\rm KM}}}

\renewcommand{\subst}{\!\leftarrow\!}

\newcommand{\Rel}{\mathrel\diamond}

\newcommand{\pl}{\prec}
\newcommand{\pg}{\succ}
\newcommand{\pdiv}{\amalg}
\newcommand{\concatrel}{\mathord\bullet}
\newcommand{\concat}{\mathord\circ}

\newcommand{\dotRel}{\stackrel{\textstyle.}{\Rel}}

\newcommand{\dotin}{\!\stackrel{\textstyle.}{\scriptstyle\in}\!}

\newcommand{\dotpl}{\stackrel{\textstyle.}{\pl}}
\newcommand{\dotpg}{\stackrel{\textstyle.}{\pg}}
\newcommand{\dotpdiv}{\stackrel{\textstyle.}{\pdiv}}

\mathchardef\orgalpha="010B
\mathchardef\orgbeta="010C
\newcommand{\VP}{\orgalpha_{\cal P}}

\newcommand{\VX}{\orgalpha_{\cal X}}

\newcommand{\PathVars}{\V_\P}

\newlength{\rulestretch}
\newlength{\rulecommentindent}
\newenvironment{rules}
    {\begingroup%
      \medskip%
     \begin{tabular}{lcl}}%
    {\end{tabular}\endgroup\medskip}

\newenvironment{rules-left}
    {\begingroup%
      \medskip%
     \begin{tabular}{lll}}%
    {\end{tabular}\endgroup\medskip}

\newenvironment{rules-left-narrow}
    {\begingroup%
      \medskip%
     \begin{tabular}{@{}l@{$\;$}ll}}%
    {\end{tabular}\endgroup\medskip}

\newenvironment{doublecol-rules}
    {\begingroup%
      \medskip%
     \begin{tabular}{lcl|lcl}}%
    {\end{tabular}\endgroup\medskip}    

\newenvironment{doublecol-rules-restr-length}[3]
    {\begingroup%
      \setlength{\tabcolsep}{#1}
      \bigskip%
     \begin{tabular}{p{#2}cl|@{\extracolsep{1.5mm}}p{#3}cl}}%
     {\end{tabular}\endgroup\bigskip}
      
\newcommand{\myrule}[3]%
    {{\rm (#1)}\rule[-0.5\rulestretch]{0pt}{\rulestretch} &%
     ${\setrefto{{\rm #1}}\ignorespaces\displaystyle #2}\over%
      {\vphantom{.}\atop\displaystyle #3}$ &}
\newcommand{\myruleindent}[4]%
    {{\rm (#2)}\rule[-0.5\rulestretch]{0pt}{\rulestretch} &%
     \hspace*{#1}   
      ${\setrefto{{\rm #2}}\ignorespaces\displaystyle #3}\over%
      {\vphantom{.}\atop\displaystyle #4}$ &}
\newcommand{\condrule}[4]%
    {{\rm (#1)}\rule[-0.5\rulestretch]{0pt}{\rulestretch} &%
     ${\setrefto{{\rm #1}}\ignorespaces\displaystyle #2}\over%
      {\vphantom{.}\atop\displaystyle #3}$ &%
     $#4$}

\newcommand{\condruleindent}[5]%
    {{\rm (#2)}\rule[-0.5\rulestretch]{0pt}{\rulestretch}  &%
     \hspace*{#1}
     ${\setrefto{{\rm #2}}\ignorespaces\displaystyle #3}\over%
      {\vphantom{.}\atop\displaystyle #4}$ &%
     $#5$}

\newlength{\ruletextwidth}
\newenvironment{rulecomment}%
  {
   \begin{minipage}[t]{0.80\ruletextwidth}}%
  {\end{minipage}}

\newenvironment{rulecomment-restr}[1]%
  {
   \begin{minipage}[t]{#1}}%
  {\end{minipage}}

\newcommand{\imodels}{\models_\I}

\begin{document}

\setlength{\ruletextwidth}{0.45\textwidth}

\title{Controlling Functional Uncertainty}
\author{Rolf Backofen\institute{Deutsches Forschungszentrum f\"ur K\"unstliche Intelligenz GmbH.}
}

\maketitle

\begin{abstract}
There have been two different methods for checking the satisfiability of
feature descriptions that use the functional uncertainty device,
namely~\cite{Kaplan:88CO} and \cite{Backofen:94JSC}. Although only the one
in \cite{Backofen:94JSC} solves the satisfiability problem completely,
both methods have their merits. But it may happen that in one single
description, there are parts where the first method is more
appropriate, and other parts 
where the second should be applied. In this paper, we present a
common framework that allows one to combine both methods. 
This is done by presenting a set of rules for simplifying feature descriptions.
The different methods are described as different controls on
this rule set, where a control specifies in which order the different
rules must be applied. 
\end{abstract} 



\section{Introduction} 

This paper is concerned with an extension to feature
descriptions, which has been introduced as ``functional
uncertainty'' in~\cite{KaplanZaenen:88,Kaplan:88CO}. This formal
device plays an important role in the framework of LFG
in modeling so-called long distance dependencies and constituent
coordination. For a detailed linguistic motivation
see~\cite{KaplanZaenen:88,KaplanZaenen:88OEGAI,Kaplan:88CO,KellerReport91}. 
Functional uncertainty consists of constraints of the form  $xLy$, where
$L$ is regular expression. $xLy$ is interpreted as
\(
         \bigvee \{ xwy \mid w \in L \}.
\)
Since this disjunction may be infinite, functional uncertainty gives
additional expressivity. Let us recall an example
from~\cite{Kaplan:88CO} and consider the topicalized
sentence {\em Mary John telephoned yesterday}. Using $s$ as a variable
denoting the whole 
sentence, the LFG-like clause $s\; topic\; x \land s\; obj\; x$ specifies
that in $s$, {\em Mary} should be interpreted as the object of the
relation {\em telephoned}.  The sentence could be extended by
introducing additional complement predicates, as e.g.\ in
sentences like {\em Mary John claimed that Bill telephoned}; {\em Mary John
claimed that Bill said that \ldots Henry telephoned yesterday}; \ldots.
For this family of sentences the  clauses
$s\; topic\; x \land s\; comp\; obj\; x$, $s\; topic\; x \land s\; comp\; comp\; obj\; x$ and
so on would be appropriate; specifying all
possibilities would yield an infinite disjunction. Using functional
uncertainty, it is possible to have a finite presentation of this
infinite specification, namely the clause $s\; topic\; x \land
s\; comp^*\, obj\; x$. 

It was shown in~\cite{Kaplan:88CO} that
consistency of feature 
descriptions is decidable, provided that
a certain acyclicity condition is met. More recently,
\cite{Backofen:94JSC} has shown that the satisfiability problem is
decidable without
additional conditions.
\nocite{baader:et:al:jlli-92}
Both algorithms have their merits. The one in \cite{Backofen:94JSC}
solves the satisfiability problem using an extended syntax, which 
makes it possible to avoid the computational explosion that causes the
undecidability in the 
cyclic case. But there are cases where the additional syntax causes some
overhead. In these cases, one would like to switch to the method used in
\cite{Kaplan:88CO}, where this overhead is avoided. On the other hand,
the algorithm in \cite{Kaplan:88CO}, which is used in the
implementation of the LFG system, cannot be extended to the
cyclic case.

In this paper, we present a new algorithm that allows one to combine
both methods under a common framework. We use the extended syntax as
proposed in~\cite{Backofen:94JSC} 
and present a new set of rewrite rules. The different methods
used in~\cite{Backofen:94JSC} and \cite{Kaplan:88CO} can then be
described as different control on this rule set, where a control
specifies the order of rule application. Thus, it
is now possible to compare both algorithms and their effects. 
In \cite{Backofen:94JSC}, this was not possible since the set of rules
presented there was tailored for the purpose of proving decidability.
As an extension, we present a control which allows the flexibility to switch
between both methods. This flexibility is needed since none of the methods
is optimal for all parts of a clause. Which
one is best depends on the regular languages used in the corresponding
part.

In Section~2, we present some needed preliminaries. In Section~3, we
introduce the input clauses and two different output clauses of our
algorithm. In Section~4 and 5 we present the rule system and some of its basic
properties. Equipped with these tools, we turn to the most interesting
part in Section~6, where we define three different controls for the
given set of rules and compare their properties.



\section{Preliminaries}

Our {\em signature} consists of a set of  {\em sorts} ${\em \cal S}$
$(A,B,\ldots)$, 
{\em first-order variables} ${\cal X}$ $(x,y,\ldots)$, {\em path
variables} $\P$ $(\myalpha,\mybeta,\ldots)$,
and {\em features} ${\F}$ $(f,g,\ldots)$. We 
assume a finite set of features and infinite sets of variables and sorts.
A {\em path} is a finite string of features.
A path $u$
is a {\em prefix} of a path $v$ (written $u \pl v$) if there is a
non-empty path 
$w$ \st\ $ v = uw$. Note that $\pl$ is neither symmetric nor reflexive.
Two paths $u,v$ {\em diverge} (written $u \pdiv v$)
if there is a common, possibly empty prefix $w$ of $u,v$ and paths
$w_1,w_2$ \st\ 
\(
        u = wfw_1
        \land
        v = wgw_2 
\)
Clearly, $\pdiv$ is a  symmetric relation.
Furthermore, for any pair of paths $u$ and $v$, then exactly one of the
relations $u = v$, $u \pl v$, $u \pg v$, or $u \pdiv v$ holds.  

A {\em simple path term} ($\pterms,\ptermt,\ldots$) is either a feature
or a path variable. A {\em path term} ($\ptermp,\ptermq,\ldots$) is
either a simple path term or a concatenation of two path terms 
$\ptermp \concat \ptermq$ (called a {\em complex path term}).
The set of constraints is given by
\[
  \begin{array}[t]{ll|ll}
     Ax & \mbox{\em sort restriction} &      \ptermp \dotpdiv \ptermq &
     \mbox{\em divergence}
 \\
     x \doteq y & \mbox{\em agreement} &     \ptermp \dotpl \ptermq &
     \mbox{\em prefix} \\ 
     x \frel{\ptermp} y & \mbox{\em subterm agreement} &
     \ptermp \doteq \ptermq & \mbox{\em path equality}\\
&&     \ptermp \dotin L & \mbox{\em path restriction} 
  \end{array}
\]
We exclude empty paths in subterm agreement since
$x\epsilon y$ is equivalent to $x\doteq y$, and use $\ptermp \dotpg
\ptermq$ as a synonym for $\ptermq \dotpl \ptermp$.
A {\em clause} is a finite set of
constraints denoting their conjunction.

An {\em interpretation}  $\I$ is a standard first-order structure, where
every feature $f\in \F$ is interpreted as a binary, functional relation
$F^\I$ and where sort symbols are interpreted as unary, disjoint
predicates (hence $A^\I \cap B^\I = \eset$ for $A\neq B$). 
A {\em valuation} is a pair $(\VX,\VP)$, where $\VX$ is a standard
first-order valuation of the variables in $X$ and $\VP$ 
is a function $\VP: \P \rightarrow \F^+$. We define $\VP(f) = f$ for
every feature $f \in \F$, and 
$\VP(\ptermp\concat\ptermq)$ to be the path $\VP(\ptermp)\VP(\ptermq)$. 
Validity for sort restrictions and agreement constraints is
defined as usual. The other constraints are valid 
in an interpretation $\I$ under a valuation $(\VX,\VP)$ iff
\[
\renewcommand{\arraystretch}{1.1}
\setlength{\arraycolsep}{1mm}
\!\!\begin{array}{lll}
  (\VX,\VP) \imodels x \frel{\ptermp} y &:\Longleftrightarrow
                                  & \alpha(\ptermp) = f_1\cdots f_n
                                  \mbox{ and } \\ &&(\VX(x), \VX(y)) \in
                                  F_1^\I \concatrel \ldots \concatrel F_n^\I\\
  (\VX,\VP) \imodels \ptermp \dotin L &:\Longleftrightarrow & \VP(\ptermp) \in L\\
  (\VX,\VP) \imodels \ptermp \dotRel \ptermq & :\Longleftrightarrow
                                    & \VP(\ptermp) \Rel \VP(\ptermq) \mbox{ for } \dotRel\; \in \{\dotpdiv,\dotpl,\doteq \}.
\end{array}
\]
where $\concatrel$ denotes binary concatenation of relations.
Note that the validity of a path constraint depends only on the path
valuation. The set of all $\X$-solutions of a clause $\phi$ in some
interpretation $\I$ is the set of all valuations $\VX$ such that there
is a path valuation $\VP$ with $(\VX,\VP) \imodels \phi$.



\section{Prime, Pre-Solved, and Solved Clauses}

In this section, we define the input and output clauses for both
phases of the algorithm. In the following, we consider only those
clauses $\phi$ such that for every distinct pair of variables $x,y$, 
$x \doteq y$ is in $\phi$ if and only if $x \neq y$ and $x$ occurs
only once in $\phi$.
A clause $\phi$ is called {\em prime} iff
\vspace*{-1\baselineskip}
{\renewcommand{\theenumi}{Pr\arabic{enumi}}
\addtolength{\leftmargin}{0.5cm} 
\begin{enumerate}
\item  every path term in $\phi$ is
       simple,
\item \label{prime-pathvar-unique} for every path variable $\myalpha$
      used in $\phi$ there is {\oldem at most\/} one constraint
      $x\frel{\myalpha} y \in \phi$, and
\item $\phi$ has no constraints of the forms
      $\pterms \dotpdiv \ptermt$, $\pterms \dotpl \ptermt$, or
      $\pterms \doteq \ptermt$. 
\end{enumerate}}
\vspace*{-1\baselineskip}

Kaplan/Maxwell~\cite{Kaplan:88CO} formulated the satisfiability problem for
functional  
uncertainty in an unsorted syntax. Essentially, this syntax consists of
the atomic constraints $Ax$, $x\, f\, y$ and $x \doteq
y$ together with the additional constraint $xLy$. Constraints of this form are
interpreted as
\( 
         x L y = \bigvee \{ xwy \mid w \in L \}.
\)
A clause $\phi$ in Kaplan/Maxwell Syntax can be translated into an
clause in our syntax by replacing every constraint $xLy$
by $x \frel{\myalpha} y \land \myalpha \dotin L$, where $\myalpha$ is a
new variable. The resulting clause will have the same $\X$-solutions.
The resulting clauses are prime clauses and hence our input clauses.
A clause is called {\em \simplified} iff
\label{simpified}
\vspace*{-1\baselineskip}
{\renewcommand{\theenumi}{Si\arabic{enumi}}
\addtolength{\leftmargin}{0.5cm} 
\begin{enumerate}
\item $Ax \in \phi$ and $Bx \in \phi$ implies $A = B$,
\item $\ptermp \dotin L \in \phi$ and $\ptermp \dotin L' \in \phi$ implies
      $L=L'$,
\item $\ptermp \dotin \eset$ is not in $\phi$,
\item $f \dotin L$ implies that $f$ is an element of denotation of $L$,
\item $x\frel{f} y\in \phi$ and $x\frel{f} z \in \phi$ implies $y=z$,
\item $\phi$ contains no constraint of the form $\pterms \doteq \ptermt$ or
      $\pterms \dotpl \ptermt$,
      \label{no-eq-or-pl}
\item every path term in $\phi$ is simple.
      \label{pre-solved-concat}
\end{enumerate}}
\vspace*{-1\baselineskip}
A \simplified\ clause is called {\em pre-solved} iff
\vspace*{-1\baselineskip}
{\renewcommand{\theenumi}{Ps\arabic{enumi}}
\addtolength{\leftmargin}{0.5cm} 
\begin{enumerate} 
\item $\pterms \dotpdiv \ptermt \in \phi$ if and only $\pterms \neq
      \ptermt$, either $\pterms$ or $\ptermt$ is a 
      path variable, and there is an $x$
      such that $\{x \frel{\pterms} y,  x \frel{\ptermt} z \}
      \subseteq \phi$. 
      \label{main-cond}
\end{enumerate}}
\vspace*{-1\baselineskip}

Pre-solved clauses are not consistent per se, since it might be that a
divergence constraint contradicts some of the path restrictions. E.g., the
pre-solved clause $x \frel{\myalpha}{y} \land x \frel{\mybeta} z \land
\myalpha \dotpdiv \mybeta \land \myalpha \dotin f^+ \land \mybeta \dotin
(ff)^+$ is inconsistent. A clause 
$\phi$ is called {\em solved} if it is either $\bot$, or it is
\simplified\ and satisfies 
\vspace*{-1\baselineskip}
\label{solved-def}
{\renewcommand{\theenumi}{So\arabic{enumi}}
\addtolength{\leftmargin}{0.5cm} 
\begin{enumerate}
\item \label{no-div} $\phi$ contains no constraint of form $\ptermp
      \dotpdiv \ptermq$, and
\item \label{free-path-variables} if $x\frel{\myalpha} y$ is in $\phi$,
      then there is no 
      $x\frel{\pterms} z$ with $\pterms \neq \myalpha$ in $\phi$.
\end{enumerate}}

\begin{lemma}
   \label{consist-normal-form}
   Let $\phi$ be a pre-solved clause different from $\bot$. Then $\phi$ is satisfiable iff
   there is a path valuation $\VP$ with $\VP \models \phi_p$, where
   $\phi_p$ is the set of constraints in $\phi$ of the forms $\pterms
   \dotpdiv \ptermt$ or $\pterms \dotin \phi$.
   \label{solved-consistency}
Furthermore, every solved clause different from $\bot$ is satisfiable.
\end{lemma}



\section{Simplification Rules}


\begin{figure*}
\fbox{
\begin{tabular}{l||l}
\begin{minipage}{0.55\textwidth}
\vspace*{-0.5cm}
\begin{doublecol-rules-restr-length}{0.5mm}{1cm}{1cm}
\condruleindent{-0.85cm}
   {Join}
   {\label{Join}\ptermp \dotin L \land \ptermp \dotin L' \land \psi}
   {\ptermp \dotin (L\cap L') \land \psi}
   {\hspace*{-0.75cm}L\neq L'} &
\myrule
   {Empty}
   {\label{Empty} \ptermp \dotin \eset \land \psi}
   {\bot} \\
\condruleindent{-0.6cm}
   {FClash}
   {\label{FClash} f \dotin L \land \psi}
   {\bot}
   {\hspace*{-0.75cm}f \not \in L} &
\condruleindent{-0.6cm}
   {SClash}
   {\label{SClash}Ax \land Bx \land \psi}
   {\bot}
   {\hspace*{-0.75cm}A \neq B} \\ \hline \hline
\myrule
   {DClash1}
   {\label{DClash1}\pterms \dotpdiv \pterms \land \psi}
   {\bot} &
\myrule
   {DClash2}
   {\label{DClash2} \pterms \concat \myalpha \dotpdiv \pterms  \land \psi}
   {\bot} \\
\myrule
   {Div1}
   {\label{Div1}\pterms \concat \myalpha \dotpdiv \pterms \concat \mybeta \land \psi}
   {\myalpha \dotpdiv \mybeta \land \psi}&
\myrule
   {Div2}
   {\label{Div2} \pterms \concat \myalpha \dotpdiv
   \mybeta \land \pterms \dotpdiv \mybeta  \land \psi}
   {\pterms \dotpdiv \mybeta \land \psi} \\
\myrule
   {DivInst}
   {\label{DivInst}\pterms\concat\myalpha \dotpdiv g \land \psi}
   {\pterms \dotpdiv g  \land \psi}&
\condruleindent{-0.75cm}
   {Triv1}
   {\label{TrivInst} f \dotpdiv g  \land  \psi}
   {\psi} 
   {\hspace*{-0.75cm}f \neq g} \\
\myrule
   {RelD}
   {\label{RelDiv}\pterms\concat\myalpha \dotpdiv \mybeta \land \psi}
   {\pterms \{\dotpdiv,\dotpl\} \mybeta \land \pterms\concat\myalpha
   \dotpdiv \mybeta \land \psi} &
\condruleindent{-0.75cm}
   {Triv2}
   {\label{TrivInst2} f \concat \pterms\dotpdiv g \concat \ptermt  \land  \psi}
   {\psi} 
   {\hspace*{-0.75cm}f \neq g} \\
& \multicolumn{2}{l|}{
        \begin{rulecomment-restr}{0.45\ruletextwidth}
        if $\pterms,\mybeta$ are unrelated
        \end{rulecomment-restr} 
}
\end{doublecol-rules-restr-length}
\end{minipage} 
&
\begin{minipage}{0.4\textwidth}
\vspace*{-0.5cm}
\begin{rules}
\myrule
    {Eq1}
    {\label{Eq1} x\frel{f} y \land  x\frel{f} z \land \psi}
    {z \doteq y \land  x\frel{f} y \land
    \psi[z\subst y]} \\
 \myrule
    {Eq2}
    {\label{Eq2}\myalpha \doteq \pterms \land x\frel{\pterms} y \land
      x\frel{\myalpha} z  \land \psi}
    {z \doteq y \land  x\frel{\pterms} y \land
    \psi[\myalpha\subst\pterms, z\subst y]} \\
\condrule
    {Pre}
    {\label{Pre}\pterms \dotpl \myalpha \land x\frel{\pterms} y \land
      x\frel{\myalpha} z \land \psi}
    {x \frel{\pterms} y \land y \frel{\myalpha} z \land 
     \psi[\myalpha \subst \pterms \concat \myalpha]}
    {\pterms\neq \myalpha} 
\end{rules}
\hrule\vspace*{0.55mm}\hrule
\begin{rules}
\myruleindent{-0.5cm}
   {DecFeat}
   {\label{DecFeat}f \concat \myalpha \dotin L \land \psi}
   {\myalpha\dotin f^{-1}L \land \psi}
\\
\condruleindent{-0.5cm}
   {DecClash}
   {\label{DecClash}\myalpha \concat \mybeta \dotin L \land \psi}
   {\bot}
   {\mbox{if $\forall w \in L: |w| = 1$}}\\
\condruleindent{-0.5cm}
   {\mbox{Dec$_\protect\decfun$}}
   {\label{LangDec}\myalpha \concat \mybeta \dotin L \land \psi}
   {\myalpha\dotin P \land \mybeta\dotin S \land \psi}
   {(P,S) \in \protect\decfun(L)}
\end{rules} 
\end{minipage}
\end{tabular}}
\caption{The simplification rules $\Rsimpl^{\decfun}$. $\protect\decfun$  associates to
every regular expression $L$ a set of decompositions $(P,S)$ with
$P\concat S \subseteq L$ }
\label{figure-simpl-rules}
\end{figure*}


The first set of rules, $\Rsimpl^{\decfun}$, is displayed in
Figure~\ref{figure-simpl-rules} and allows one to simplify a clause
satisfying certain restrictions that will be captured under the notion
of a admissible clause.
Most of the rules are deterministic, i.e., replacing a clause with
the result of applying one of these rules yields a clause having the
same $\X$-solutions. The rules $(\ref{RelDiv})$ and
$(\ref{LangDec})$ are non-deterministic rules, which implies that we
have to replace a clause by the disjunction of all possible applications
of the corresponding rule. Thus, applying  $(\ref{RelDiv})$ to a clause
of the form
\(
\myalpha\concat \myalpha \dotpdiv \mybeta \land \ldots
\)
yields the disjunction 
\[
(\myalpha\concat \myalpha' \dotpdiv \mybeta \land \myalpha \dotpdiv
\mybeta \land  \ldots) \ \lor \ (\myalpha\concat \myalpha' \dotpdiv \mybeta \land \myalpha \dotpl
\mybeta \land  \ldots)
\]

{\renewcommand{\A}{{\cal A}}
The rule set is indexed by the decomposition function $\decfun$ used in
$(\ref{LangDec})$. The simplest version of $\decfun$ just decomposes a
regular language $L$ into a set of pairs $(P,S)$ with the property that
there is a state $q$ in the minimal automaton $\A$ for $L$ with $P = \{ w
\neq \epsilon\mid \delta_\A(q_{\rm in},w) = q \}$ and $S= \{ w \neq
\epsilon \mid \delta_\A(q,w) \in {\rm Fin}_\A\}$. Here, $q_{\rm in}$ is
the initial state, ${\rm Fin}_\A$ is the set of final states and
$\delta_\A$ the transition function of $\A$.} This decomposition
function is sufficient for the case of non-cyclic clauses. For cyclic
clauses, we have to use a different  decomposition function (as will
explained later). In any case, in order to preserve all solutions of a
clause the decomposition function has to satisfy 
\[
   \begin{array}{l}
        \forall L,\; \forall w_1,w_2 \neq \epsilon:\\
        \;\;\;\;\;\;
             [w_1w_2\in L 
              \impl
              \exists P,S \in\decfun(L): (w_1 \in P \land w_2 \in S)].
   \end{array}                
\]

The simplification does not handle arbitrary clauses. E.g.,
we handle only those prefix and equality constraints $\pterms
\dotpl \ptermt$ and $\pterms \doteq \ptermt$ in a clause $\phi$ with
the property that 
there is a variable $x$ and variables $y,z$ such that $x
\frel{\pterms} y$ and $x \frel{\ptermt} z$ is in $\phi$. Furthermore,
the rules cannot reduce divergence constraints of the form $\pterms \concat
\pterms' \dotpdiv \ptermt \concat \ptermt'$ with $\pterms \neq \ptermt$,
and the control imposed on our rewrite rules carefully avoid such
constraints.
The reason is that for decomposing  
the complex path terms in  $\pterms \concat
\pterms' \dotpdiv \ptermt \concat \ptermt'$, we might be forced to
introduce complex path terms that have a length greater than $2$, which
we must avoid to achieve a quasi-terminating rewrite system.

We now define the restriction imposed on derivable clauses.
Given a clause $\phi$, we define the {\em outgoing edges} of a
first-order variable $x$ in $\phi$ as
\begin{eqnarray*}
\outgoing{\phi}{x} & := \{ \pterms \mid \mbox{there is $z$ with
$x\frel{\pterms} z \in \phi$} \}
\end{eqnarray*} 
We say that a variable $x$ in $\phi$ is {\em tagged} if there is a
prefix constraint $\pterms \dotpl \myalpha$ in $\phi$ with
$\{\pterms,\myalpha\} \subseteq \outgoing{\phi}{x}$.
A clause is is called {\em admissible} if 
$\phi$ contains no complex path terms in prefix or path equality
constraints and
\vspace*{-1\baselineskip}
{\renewcommand{\theenumi}{Ad\arabic{enumi}}
\addtolength{\leftmargin}{0.5cm}
\begin{enumerate}
\item \label{basic-pathvar-unique} for every path variable $\myalpha \in
      \PathVars(\phi)$, there is {\oldem exactly} one constraint $x\frel{\myalpha} y \in \phi$,
\item \label{local-relation} for every path constraint of the forms
      $\pterms \{\doteq,\dotpl,\dotpdiv \} \ptermt$ in $\phi$, there exists
      a variable $x$  such that $\{\pterms,\ptermt \} \subseteq
      \outgoing{\phi}{x}$,
\item \label{prefix-no-eq} 
      if $\phi$ contains a prefix constraint, then $\phi$ contains no
      path equality constraint,
\item \label{tagged-var} if $\phi$ contains {\oldem at most\/} one
      tagged variable,
\item \label{diff-prefix} if $\phi$ contains two different prefix
      constraints $\pterms 
      \dotpl \myalpha$ and $\ptermt \dotpl \mybeta$, then either
      $\pterms = \ptermt$, or $\pterms$ and $\ptermt$ are different
      features,
\item \label{no-trivial} $\phi$ contains no trivial constraints of the
      form $\pterms \dotpl \pterms$, $\pterms \dotpl f$, $f \doteq g$, or
      $f \doteq f$.
\end{enumerate}
}
The last condition just lists constraints which either are inconsistent
or superfluous. We could also get rid of these constraints using some
appropriate rewrite rules, but we think that it is more efficient to
avoid these constraints. Note that every prime clause is admissible. 
A clause is called {\em basic} if is derivable using $\Rsimpl^{\decfun}$ 
from an admissible $\phi$ that contains no complex path terms.

\begin{proposition}
\label{simpl-basic}
Every basic clause is admissible.
\end{proposition} 

The tedious part of the proof of this proposition are
the rules $(\ref{Pre})$, $(\ref{Div1})$, and
$(\ref{RelDiv})$, since one has to check whether the new introduced
constraints satisfy the conditions~\ref{local-relation} and
\ref{diff-prefix}. For this purpose, one has to record
exactly all possible effects that the introduction of complex path terms
in the $(\ref{Pre})$ rule can have on admissible clauses. E.g., it is
guaranteed by the definition of $(\ref{Pre})$ that if a basic $\phi$ contains a
complex path term $\alpha \concat \mybeta$, then
there are variables
$x,y,z$ such that $x\frel{\myalpha} y$ and $y \frel{\mybeta} z$ are in
$\phi$. This, together with condition~\ref{basic-pathvar-unique}, implies
that if $\phi$ contains  a constraint of the form $\myalpha \concat \mybeta
\dotpdiv \myalpha \concat \mybeta'$ in $\phi$, then there are variables
$x,y,z,z'$ such that $\{x \frel{\myalpha} y,y \frel{\mybeta} z , y
\frel{\mybeta'} z' \} \subseteq \phi$. Hence, we know for the new
relation introduced in~\ref{Div1}, the condition~\ref{local-relation} is
satisfied.

\begin{lemma}[Termination,Completeness]
\label{simpl-term}
A basic clause is irreducible w.r.t. $\Rsimpl^\decfun$
iff  $\phi$ is \simplified. Furthermore, for every admissible clause $\phi$ 
there are no infinite derivations starting with $\phi$, and $\phi$ has the same
$\X$-solutions as the set of \simplified\ clauses derivable from $\phi$. 
\end{lemma} 
\begin{proofsketch}
We consider only the claim of termination. Here, the $(\ref{RelDiv})$
rule is the most 
difficult part since it introduces a new relation. All other rules
reduces either the number of variables, constraints or complex path
terms. To show that
$(\ref{RelDiv})$ terminates, it is necessary to know that there is
exactly one variable $x$ in $\phi$ such that for all constraints of the
form $\pterms \concat \myalpha \dotpdiv \mybeta$ in $\phi$, both
$\pterms$ and $\mybeta$ are in $\outgoing{\phi}{x}$. But this is an
immediate consequence of the fact that there is at most one tagged
variable in $\phi$ (Condition~\ref{tagged-var}) together with Condition~\ref{local-relation}. Hence, $(\ref{RelDiv})$
only adds constraints between unrelated simple terms that are in
$\outgoing{\phi}{x}$, where $x$ is the tagged variable of $\phi$. Since
there are only finitely many possible relations, and since both the
$(\ref{Pre})$ rule and the $(\ref{Div2})$ rule do not increase the
number of unrelated simple terms in $\outgoing{\phi}{x}$, $(\ref{RelDiv})$ cannot cause non-termination. This
leads to the following termination ordering. Let $x$ be the tagged
variable in $\phi$, and let $\Thetasimpl(\phi)$ be the quadruple
\[
\begin{array}{r@{}cl}
& ( & \mbox{\#unrelated terms in $\outgoing{\phi}{x}$},\ \mbox{\#constraints},\\
& & \mbox{\#complex path terms in $\phi$},\ \mbox{\#variables})
\end{array}
\]
Then for every $r \in \Rsimpl^{\decfun}$, if $\phi'$ is the result of
applying $r$ to a basic clause $\phi$, then $\Thetasimpl(\phi) >_4
\Thetasimpl(\phi')$, where $>_4$ is the lexicographic greater ordering on quadruples.
\end{proofsketch} 

\section{Generating pre-solved and solved clauses}

As we have explained in the introduction, one of the main tools for
solving prime clauses is to ``guess'' the different relations between
path variables, and to check this
relation for consistency with the rest of the clause afterwards.
Clearly, one has to ``guess'' all possible relations, which implies that
the rules for introducing this relation must be non-deterministic.
We have already encountered one rule for non-deterministically
introducing relations between simple path terms, namely
the rule $(\ref{RelDiv})$. The other two rules are listed below and form
the rule set $\Rpre$.

\begin{rules}
\myrule
   {Relate1}
   {\label{Relate1}x \frel{\myalpha} y \land x \frel{\mybeta} z \land \psi}
   {\myalpha \{\doteq,\dotpg,\dotpl,\dotpdiv\}\mybeta \land x
   \frel{\myalpha} y \land x \frel{\mybeta} z \land \psi}\\
   & \multicolumn{2}{l}{
       \begin{rulecomment}
       $\myalpha, \mybeta$ unrelated in $\psi$, $(\ref{RelDiv})$ not
       applicable        
       \end{rulecomment}}\\
\myrule
   {Relate2}
   {\label{Relate2}x \frel{f} y \land x \frel{\myalpha} z \land \psi}
   {f \{\doteq,\dotpl,\dotpdiv\}\myalpha \land x \frel{f} y
   \land x \frel{\myalpha} z \land \psi} \\
   & \multicolumn{2}{l}{
       \begin{rulecomment}
       $f,\myalpha$ unrelated in $\psi$, $(\ref{RelDiv})$ not applicable
       \end{rulecomment}}\\
\end{rules} 

Using the following set of rules $\Rsolve$, we can transform a pre-solved clause into an
equivalent set of solved clauses. 
\begin{rules-left-narrow}
\condrule
   {Inst}
   {\label{SolveInst}\myalpha \dotpdiv f \land \psi}
   {g \dotpl \myalpha \land \myalpha \dotpdiv f \land \psi} 
   {\hspace*{-2.5cm}f \neq g}\\
\condrule
   {Intro}
   {\label{myintro} f \dotpl \myalpha \land x \frel{\myalpha} y \land \phi}
   {f \dotpl \myalpha \land  x
   \frel{\myalpha} y  \land x \frel{f} y' \land \phi}
   {\hspace*{-2.5cm}\mbox{if }\forall z: x \frel{f} z \not\in \phi}\\
\condrule
   {Solv1}
   {\label{Solve1}
    \myalpha \dotpdiv \mybeta  \land \psi}
   {f \dotpl \myalpha \land g \dotpl \mybeta 
    \land \myalpha \dotpdiv \mybeta \land \psi}
   {\hspace*{-2.5cm}f \neq g}\\
\myrule
   {Solv2}
   {\label{Solve2}\myalpha \dotpdiv \mybeta \land x \frel{\myalpha} y \land x
   \frel{\mybeta} z \land \psi}
   {f \dotpl \myalpha \land g \dotpl \mybeta 
    \land \myalpha \dotpdiv \mybeta \land 
    x \frel{\mydelta} u \land u \frel{\mybeta} y \land u \frel{\mybeta} y 
   \land \psi_{\rm subs}}\\
& 
\multicolumn{2}{l}{
\begin{rulecomment-restr}{0.4\textwidth}
\rule{0mm}{0.5cm}
where $\psi_{\rm subs} = \psi[\myalpha \subst \mydelta\concat\myalpha, \mybeta \subst
   \mydelta\concat\mybeta]$, $f \neq g$ and $\delta$, $u$ are new
variables
\end{rulecomment-restr}} 
\end{rules-left-narrow} 

The two rules~$(\ref{Solve1})$ and~$(\ref{Solve1})$ together will be
seen as one complex, non-deterministic rule called (Solve){\setrefto{{\rm
Solve}}\label{Solve}}. The $(\ref{Solve})$ directly expands a divergence
constraints into its definition, thus solving a single divergence
constraint. The $(\ref{Solve1})$ rules 
reflects the case that two paths diverge with an empty prefix while
$(\ref{Solve2})$ reflects the case that the common prefix is not empty.
Since the valuations always associates non-empty paths to path variables,
we have to distinguish these cases. Note that $(\ref{myintro})$ is the
only deterministic rule, and that all of the other rules are non-deterministic.

\begin{proposition}
If a \simplified\ clause is not
pre-solved, then one of $(\ref{Relate1})$ or $(\ref{Relate2})$  is applicable.
Furthermore, a clause is pre-solved if none of the rules in
$\Rsimpl^{\decfun} \cup \Rpre$ is applicable, and solved if none of the
rules in $\Rsimpl^{\decfun} \cup \Rpre \cup \Rsolve$ is applicable.
\end{proposition} 

\section{Controlling rule application}

In this section, we present different possible controls over the set of
rules given by $\R^{\decfun} = \Rsimpl^{\decfun} \cup \Rpre \cup
\Rsolve$. A {\em control} is a partial order $\controlconst{\control}$ on
$\R^{\decfun}$. A 
derivation $\phi_1 \rightarrow_{r_1} \phi_2 \ldots$ is {\em licensed}
by a control $\controlconst{\control}$ iff for every step $\phi_i
\rightarrow_{r_i} 
\phi_{i+1}$, no rule instance $r$ with $r \controlconst{\control} r_i$ is
applicable. We use $\controlconst{\control}$-derivative and
$\controlconst{\control}$-derivation in the obvious way.

If we would apply the rules without any control, then not only
is termination not guaranteed, but we may even produce a clause that is
not admissible. E.g.,  consider the clause 
\(
        x \frel{\myalpha} y \land x\frel{\mybeta} z \land x
        \frel{\mybeta'} z' .
\)
Then applying $(\ref{Relate1})$ twice may produce the clause
\(
        \myalpha \dotpl \mybeta \land \mybeta \dotpl \mybeta' \land x \frel{\myalpha} y \land x\frel{\mybeta} z \land x
        \frel{\mybeta'} z',
\)
which is not admissible since it does not fulfill condition~\ref{diff-prefix}.
Hence, our minimal control $\basiccon$ guarantees that
the simplification rules are applied before one of the rules
in $\Rpre \cup \Rsolve$ are applied, i.e.,
\[
        \forall r \in \Rsimpl^{\decfun}, \forall r' \in\Rpre \cup
        \Rsolve: r \basiccon r'.
\]

\begin{proposition}
If $\phi$ is derivable with $\R^{\decfun}$ from a prime clause using the
control $\basiccon$, then $\phi$ is admissible.
\end{proposition} 

\begin{proofsketch}
This follows from the fact that if $\phi$ is an admissible clause that
contains no complex path terms (which prime clauses are), then it is
basic and therefore admissible due to Proposition~\ref{simpl-basic}.
Furthermore, it can be simplified due to Lemma~\ref{simpl-term}.
Hence, according to the control $\basiccon$, we can apply a rule in
$\Rpre\cup\Rsolve$ if and only if the corresponding rule is simplified. 
And it is easy to check that applying a rule $\Rpre\cup\Rsolve$ to a
simplified clause yields an admissible clause that contains no complex
path terms.
\end{proofsketch} 

If for every prime clause $\phi$ there are no infinite derivations using
$\R^{\decfun}$, then we know that we could transform every prime clause
$\phi$ into an equivalent set of solved clauses. But this is not the
case. Consider e.g. the clause
\[
        x \frel{\myalpha} x \land x \frel{f} y \land \myalpha \dotin f^+
        \land Ax \land By.
\]
Then applying $(\ref{Relate2})$ to introduce a
constraint $f \dotpl \myalpha$ followed by an application of 
$(\ref{Pre})$ and $(\ref{DecFeat})$ yields the same clause again. The
reason for the loop is that we have a cyclic description of the form $x
\frel{\myalpha} x$. But we can show that, similar to Kaplan/Maxwell's
Algorithm, $\R^{\decfun}$ is terminating under $\basiccon$
if no cyclic descriptions are encountered.

\begin{theorem}
\label{non-cycle-term}
Let $\phi$ be a prime clause such that no $\basiccon$-derivative of
$\phi$ contains a cycle. Then there is no infinite
$\basiccon$-derivation. Furthermore, $\phi$ has the same $\X$-solutions
as the set of solved clauses derivable from $\phi$.
\end{theorem} 

Hence, the control $\basiccon$ can be used if one does not want to handle
cyclic structures. Note that one can easily recognize whether the
algorithm runs in a loop using an occurs check (i.e., by checking
whether one visits 
some variable twice). In this case, one can
either stop (without knowing anything about the satisfiability), or
switch to the more complex control $\quasicon$ that at least guarantees
quasi-termination. A
rewrite system is quasi-terminating, if it may loop, but produces
only finitely many different clauses. 
Given a quasi-terminating rewrite
system, an algorithm using this system must record the previously
calculated clauses and stop, if one clause is produced for the second
time. This is expensive, but necessary if you want to handle cyclic
structures. $\quasicon$ is the control extending $\basiccon$ with the
property that 
\[ 
   \forall r \in \Rpre, \forall r' \in \Rsolve: r \quasicon r'.
\]
 Since in $\quasicon$ the rules in $\Rpre$ are applied first, we
know that every clause is first transformed into a set of pre-solved
clauses, which are then solved using $\Rsolve$. By an adaptation
of~\cite{Backofen:94JSC} we get the following theorem.

A necessary condition for this
theorem is that for every prime clause $\phi$, the set
of all regular languages introduced in some $\quasicon$-derivative of
$\phi$ by $(\ref{LangDec})$, $(\ref{DecFeat})$, or $(\ref{Join})$
is finite. Clearly, there are only finitely many different  regular
languages produced by $(\ref{DecFeat})$ or $(\ref{Join})$, but
$(\ref{LangDec})$ may be a problem. \cite{Backofen:94JSC} shows how an
appropriate decomposition function can be found for a given
prime clause $\phi$.

\begin{theorem}
\label{quasi-term}
There exists a decomposition function $\decfun$ such that for every
prime clause $\phi$ there are only finitely many 
$\quasicon$-derivatives. Furthermore, $\phi$ has the same $\X$-solutions
as the set of $\quasicon$-derivatives that are solved.
\end{theorem}

Next, we want to show that we can simulate Kaplan/Maxwell's algorithm.
The idea is that one can associate with every application of one of the
$\Rpre$ rules a corresponding rule in Kaplan/Maxwell's algorithm. But
in
their algorithm, there is no syntactic equivalent for the prefix,
divergence, and path equality constraints. The path equality and prefix
constraints are not a problem, since they will be removed under $\basiccon$
before the next rule in $\Rpre$ is applied. But the divergence
constraints may survive. This is not the case if we apply the rules in
$\Rsolve$ before applying a rule in $\Rpre$. Hence, we can define the
control $\maxwell$ extending $\basiccon$ by
\[      
   \forall r \in \Rsolve, \forall r' \in \Rpre: r \maxwell r'.
\]
Note that this control orders the rules in $\Rpre$ and $\Rsolve$ in
exactly the other direction than the control $\quasicon$. The control
$\maxwell$ has the property that there are no divergence constraints in
the derivable, simplified clauses, which implies that we can translate
them back into the Kaplan/Maxwell syntax. 
Furthermore, the only rule for
handling the divergence constraints that is needed under this control is
$(\ref{TrivInst})$, i.e., the handling of divergence constraints under this
control is trivial. 

\begin{theorem}
Let $\phi$ be a prime clause and let $\phi'$ be a $\maxwell$-derivative.
Then a rule in $\Rpre$ is applicable if and only if $\phi'$ is
simplified and contains no divergence constraints. Furthermore, 
we can associate with every $\maxwell$-derivation $D$ a corresponding
derivation $D'$ in Kaplan/Maxwell's algorithm such that the sequence of
translations of clauses in $D'$ is exactly the sequence of
simplified clauses in $D$, and vice versa. 
\end{theorem} 

Now what's left? This are the $\basiccon$-derivations that are neither
$\quasicon$-derivations nor $\maxwell$-derivations. The question arise
whether there is any use for such derivations, and there are. The
reason simply is that it depends on the used regular languages whether
for a specific divergence constraint, it is more useful to solve this
divergence constraint immediately (as it is done under the $\maxwell$
control), or whether it is better to delay this solving (as in the
$\quasicon$ control) hoping that this might be superfluous since other
rules may detect a simple inconsistence. Consider a generalization of
the example given in the introduction using regular languages of
the form $comp^+\{grel_1,\ldots,grel_n\}$, where $grel_1,\ldots,grel_n$ are
grammatical relations such as direct object, indirect object and so on.
Now let $\phi$ be a clause of the form
\[
\begin{array}{r@{\ }l@{\ }}
        x \frel{\myalpha} y \land x \frel{\mybeta} z  \land \myalpha
        \dotpdiv \mybeta \land  & \myalpha
        \dotin comp^+\{grel_1,\ldots,grel_n\} \\
        \land & \mybeta \dotin
        comp^+\{grel_1,\ldots,grel_n\} \land\psi
\end{array} 
\]
Then we know that $\myalpha$ is of the form $\mydelta \concat f \concat
\myalpha'$ and $\mybeta$ is of the form $\mydelta \concat g \concat
\mybeta'$ such that the common prefix $\mydelta$ is in $comp^+$, $f \in
\{comp, grel_1,\ldots,grel_n\}$ and $g \in \{comp,
grel_1,\ldots,grel_n\} - \{f\}$. Hence, there are $(n+1)\times n$
different possibilities that $\myalpha$ and $\mybeta$ diverge.
Solving the divergence constraint immediately as forced by the $\maxwell$
control, would produce a disjunction of $(n+1)\times n$ clauses. This
makes sense for $n=1$ (as in the case of $comp^+subj$) since it reduces
the overhead for keeping the divergence constraint. But should
be delayed in the case where $n$ is greater than $1$. Using the control
$\basiccon$, one has the flexibility to do so, and
Theorem~\ref{non-cycle-term} guarantees that the algorithm terminates in
the case of non-cyclic descriptions.


\bibliographystyle{plain}
\bibliography{abbrev,backofen-crossrefs,backofen-pub,backofen-here,backofen-other}

\end{document}